\documentclass[conference]{IEEEtran}

\usepackage[dvips]{graphics,graphicx,epsfig}
\usepackage{epsf,psfig,times,amsfonts,amsmath,algorithmic,algorithm}

\IEEEoverridecommandlockouts

\begin{document}

\title{{\ \\ \LARGE\bf Evolving Dynamic Change and Exchange of Genotype Encoding in Genetic Algorithms for Difficult Optimization Problems}}

\author{\begin{tabular}{cccc}
Maroun \textsc{Bercachi} & Philippe \textsc{Collard} & Manuel \textsc{Clergue} & Sebastien \textsc{Verel} \\
bercachi@i3s.unice.fr & philippe.collard@i3s.unice.fr & clergue@i3s.unice.fr & verel@i3s.unice.fr \\ \\
\multicolumn{4}{c}{Techniques pour l'Evolution Artificielle team (TEA)} \\
\multicolumn{4}{c}{Laboratoire I3S - Universite De Nice-Sophia Antipolis / CNRS} \\
\multicolumn{4}{c}{Les Algorithmes, 2000 Route Des Lucioles} \\
\multicolumn{4}{c}{BAT. Euclide B - BP. 121 - 06903 Sophia Antipolis - France} \\
\multicolumn{4}{c}{http://twiki.i3s.unice.fr/twiki/bin/view/TEA/WebHome} \\
\end{tabular}}

\maketitle

\begin{abstract}
The application of genetic algorithms (GAs) to many optimization problems in organizations often results in good performance and high quality solutions. For successful and efficient use of GAs, it is not enough to simply apply simple GAs (SGAs). In addition, it is necessary to find a proper representation for the problem and to develop appropriate search operators that fit well to the properties of the genotype encoding. The representation must at least be able to encode all possible solutions of an optimization problem, and genetic operators such as crossover and mutation should be applicable to it. In this paper, serial alternation strategies between two codings are formulated in the framework of dynamic change of genotype encoding in GAs for function optimization. Likewise, a new variant of GAs for difficult optimization problems denoted {\it Split-and-Merge} \textnormal {GA} ({\it SM-GA}) is developed using a parallel implementation of an SGA and evolving a dynamic exchange of individual representation in the context of Dual Coding concept. Numerical experiments show that the evolved {\it SM-GA} significantly outperforms an SGA with static single coding.
\end{abstract}

\section{Introduction}
Genetic algorithms (GAs) are search procedures based on principles derived from the dynamics of natural population genetics. These algorithms abstract some of the mechanisms found in evolution for use in searching for optimal solutions within complex "fitness landscapes" \cite{r1}. Like the natural world, GAs are forms of adaptive systems in which various chromosomes interact via sufficiently complicated elements \cite{r4}. These elements include selection method, crossover and mutation operators, the encoding mechanism ("representation") of the problem, and many others. All of these are typically preset by the user before the actual operation of a GA begins. Many individual representations have been proposed and tested within a wide-range of evolutionary models. Maybe, an essential natural question that has to be answered in all these evolutionary models : which is the optimal genotype encoding needed to make individuals evolve better in a GA application ? To prevent approximately a bad choice of a coding that do not match to a problem fitness function, the research effort reported in this paper focused on developing strategies of sequential and parallel implementations of a simple GA (SGA) evolving the use of two codings simultaneously, having the goal to revise GAs behaviour, to probably enhance GAs performances degree, and later to refine GAs solution quality.

Some previous works \cite{r10} proposed to use dynamic representations to escape local optima. Their strategies focused on parameter optimization and consisted in switching the gray representation of individuals when state-of-the-art GA has converged. In this paper, we explore different ways and diverse criteria of conversation and interaction between two representations in one SGA. The first way changes sequentially two codings according to a specific touchstone and the second way exploits in parallel two codings in a self-propelled mechanism. So currently, we focus more on the matter of exploring variant strategies of dynamic representation and we concentrate well on the topic of enhancing the basic operations and intensifying the main performances of an SGA.

The structure of our present work as follows : Section \ref{secIR} introduces individual representation character. Section \ref{secSDCS} presents a quick study about the hypothesis of GAs twofold representation in the form of diverse serial alternation strategies and section \ref{secSMGAT} presents a new technique for a {\it Split-and-Merge} GA ({\it SM-GA}) as a parallel implementation of an SGA  in the context of symmetric Dual Coding basic scheme. Section \ref{secSE} introduces the protocol of our experiments including the functions utilized to test the suggested algorithms, the set of parameters used, the numerical results of our observations and the t-test results for a later judgement. Finally, section \ref{secGDC} presents some general discussions and conclusions.

\section {Individual Representation}
\label{secIR}
Representation is one of the key decisions to be made when applying a GA to a problem. How a problem is represented in a GA individual determines the shape of the solution space that a GA must search \cite{r11}. For example, the choice of tree representation instead of vector representation could help according to the tested problem \cite{r12}. For any function, there are multiple representations which make optimization trivial \cite{r13}. However, the ensemble of all possible representations is a larger search space than that of the function being optimized. Unfortunately, practitioners often report substantial different performances of GAs by simply changing the used representation. The difficulty of a specific problem , and with it the performance of GAs, can be modified dramatically by using various types of encodings. Indeed, an encoding can perform well for many diverse test functions, but fails for the one problem which one really wants to solve \cite{r2}. These observations were confirmed by empirical and theoretical investigations. In a particular way, there are kinds of GAs, like the messy GA developed by Goldberg et al. (1989), that use an adaptive encoding that adjusts the structure of the representation to the properties of the problem. This approach, however, burdens the GA not only with the search for promising solutions, but also the search for a good representation. Generally and appending to some major studies, the use of Gray Coding (GC) has been found to enhance the performance of genetic search in some cases \cite{r8}. However, GC produces a different function mapping that may have fewer local optima and different relative hyperplane relationships than the Standard binary Coding (SC) which sometimes has been found to complicate the search for the optimum by the fact of producing a large number of local optima \cite{r9}. Also, GC has been shown to change the number of local optima in the search space because two successive real Gray-Coded numbers differ only by one bit. Moreover, the use of GC is based on the belief that changes introduced by mutation do not have such a disruptive effect on the chromosome as when we use SC \cite{r7}. Besides, we should mention that SC also seems to be effective for some classes of problems because its advantage resides by the fact that it frequently locates the optimal solution. Also, with SC the best fitness tendency to approach the global optimum is very high due to its power in discovering the search space and owing to its convergence speed to the best solution \cite{r6}. As a result, different encodings of the same problem are essentially different problems for a GA. Selecting a representation that correlates with a problem's fitness function can make that problem much easier for a GA to solve \cite{r8}. An interesting approach consists of incorporating good concepts about encodings and developing abstractive models which describe the influence of representations on measurements of GA performance. After that, dynamic representation strategies can be used efficiently in a theory-guided manner to achieve significant advancement over existing GAs for certain classes of optimisation problems.

\section{Serial Dual Coding Strategies}
\label{secSDCS}
Many optimization problems can be encoded by a variety of different representations. In addition to binary and continuous string encodings, a large number of other, often problem-specific representations have been proposed over the last few years.. As no theory of representations exists, the current design of proper representations is not based on theory, but more a result of black art \cite{r2}. Although, designing a new dynamic appropriate representation will not remain the black of art of GAs research but become a well predictable engineering task. In our study, we used to encode minimization test problems with binary strings and we referred specifically to the two most popular codings SC and GC. As has been discussed, SC has a very high tendency to converge to a local optima speedily while GC has the potential to significantly alter the number of local optima in the search space \cite{r7}. Therefore, the difficulty and the essential work is to discover the best strategy of alternance between SC and GC in order to improve GAs performances. At first, we studied the possibility of GAs dual chromosomal encryption using sequential alternation strategies such as {\it Periodic-GA}, {\it APeriodic-GA}, {\it LocalOpt-GA}, {\it HomogPop-GA} and {\it SteadyGen-GA}.

The idea for the {\it Periodic-GA} was to alternate between two given codings for the same number of generations ($period$). The parameter requires fine tuning for a given problem.

{\it Aperiodic-GA} differs from {\it Periodic-GA} by selecting, before each alternance, an arbitrary period ($aperiod$) from [$minP:maxP$] interval. The parameter does not demand expensive tuning because an interval accommodation is more easier and less sensitive while moving from one test function to another.

\begin{figure}
\begin{tabular}{@{}c@{}c@{}c@{}}
\epsfig{figure=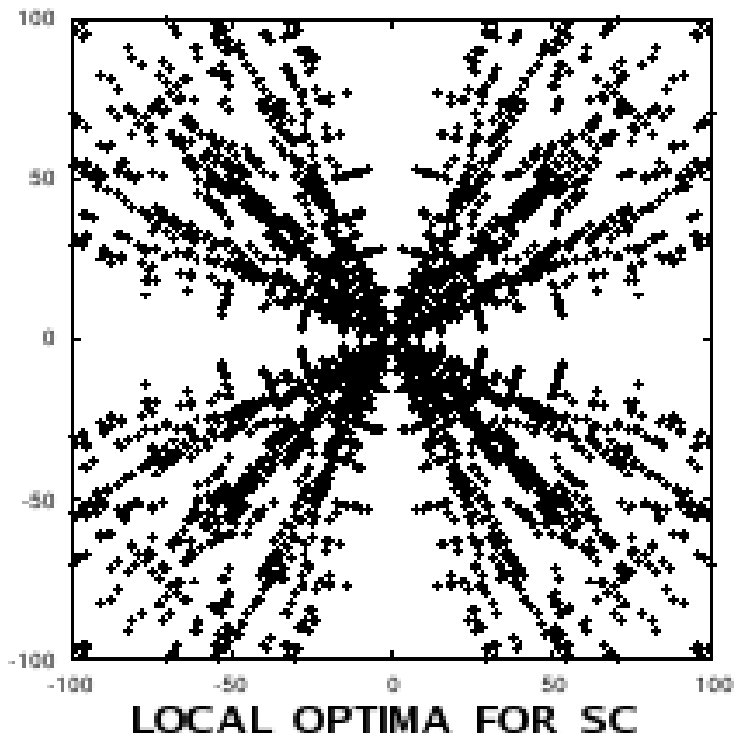, width=29mm, height=32mm} & \epsfig{figure=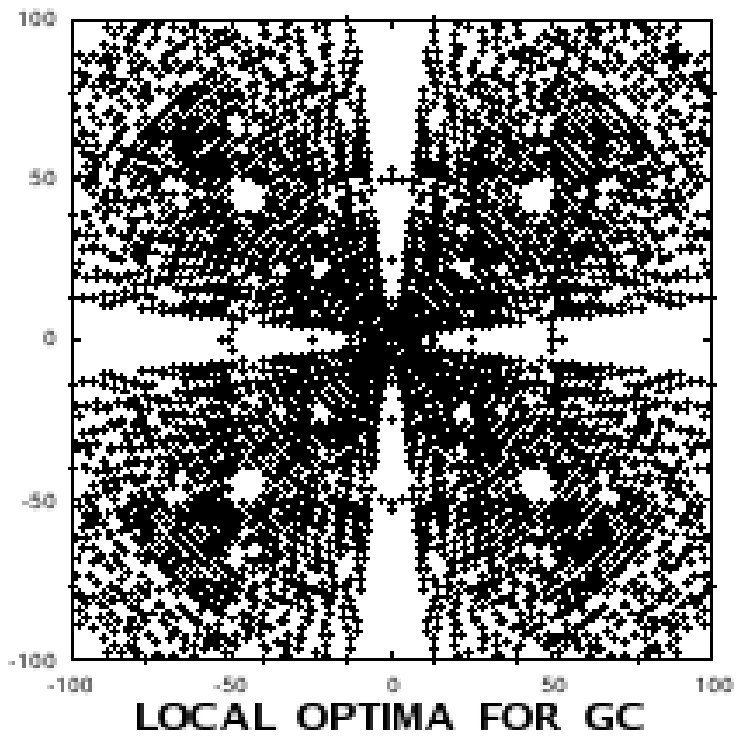, width=29mm, height=32mm} & \epsfig{figure=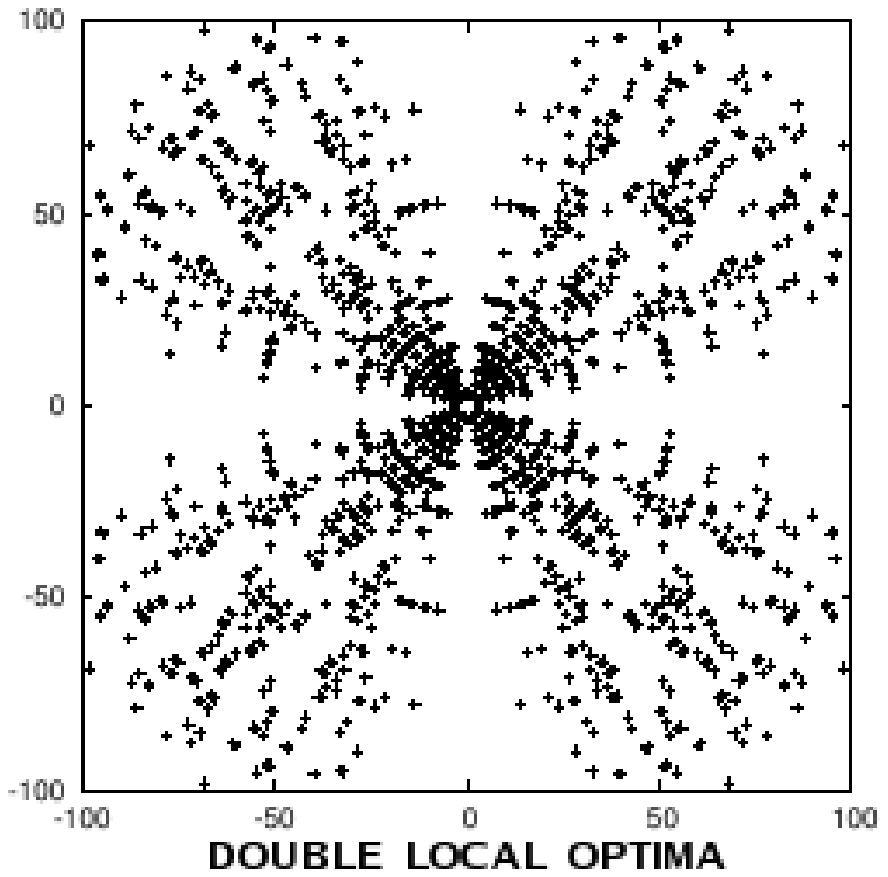, width=29mm, height=32mm}
\end{tabular}
\caption{Positions of Local Optima.}
\label{figPosLocOpt}
\end{figure}

{\it LocalOpt-GA} consists in changing coding when the population's best individual is a local optima. The idea was to try alternating between representations because a local optima under a coding is not necessary a local optima under the other, a fact that probably will permit to escape the obstacle created by a local optima and to achieve more better results. This proposition does not require any parameter and need any adjustment but it increases significantly the execution time by the fact of processing a huge number of function evaluations at each generation. In the framework of {\it LocalOpt-GA}, we studied the position and the number of local optima, for Schaffer function F6 (cf. section \ref{secTP}) and for the two codings SC and GC, by an exhaustive exploration of the search space. A double local optima is a solution which is a local optima under two used codings. For function F6, the reported number of local optima for SC was $6652$ and for GC was $7512$. Thus, there was less double local optima shared between SC and GC and the reported number was $2048$. The positions $(x, y)$ of local optima are given in Fig \ref{figPosLocOpt}.

The idea for the {\it HomogPop-GA} was to change representation when a population attains an homogeneous phase that reveals its inability to enhance more the results. Homogeneity criteria was measured by the standard deviation of fitnesses in the population in comparison with a given real number ($\varepsilon$). Also by alternating, this will keep some degree of diversity between individuals which will help in discovering the search space. The parameter is very sensitive and requires to be tuned for each problem.

In {\it SteadyGen-GA}, the alternance is realized when the best fitness value is not modified for a given number of generations ($steadyGen$), and this to keep enhanced the fitness capacity during the search. The parameter is not sensitive while tuning for a given problem.

The common main algorithm to these strategies consists in executing an SGA for one generation with a given coding. After that, it consists in testing proposal particular condition; if true, then it belongs to alternating to the other coding and converting individuals representation to that coding. Alternance cycle continues until a given maximum number of generations $maxGen$ are reached. To simplify serial strategies algorithms, common procedures were used. For a given population $pop$, given representations $coding$, $coding1$ and $coding2$, and given numbers $steadyGen$ and $maxGen$, these procedures can be resumed as follows :
\begin{itemize}
\item {{\bf Generate\_Initial\_Population}() : generates randomly an initial population.}
\item {{\bf Run\_1\_SGA}($pop$, $coding$) : executes an SGA for one generation with $pop$ having $coding$ as representation.}
\item {{\bf Alternate\_Coding}($coding1$, $coding2$) : switches problem encoding between $coding1$ and $coding2$ and returns the coding corresponding to the last altered coding.}
\item {{\bf Convert\_Population}($pop$, $coding$) : converts $pop$ individuals representation to $coding$.}
\item {{\bf Is\_MaxGen}($maxGen$) : a boolean function that returns true if an algorithm was executed entirely for $maxGen$ generations and false otherwise.}
\end{itemize}
Serial Dual Coding proposals can be defined as follows :
\begin{enumerate}
\item {\textit{Periodic-GA} (cf. Algo 1) : The alternance is realized if an SGA was executed for $period$ generations with a given coding. A boolean procedure {\bf Is\_Period}($period$) is used and returns true if an SGA was operated for $period$ generations and false otherwise.}
\item {\textit{Aperiodic-GA} (cf. Algo 2) : Same as \textit{Periodic-GA} with an arbitrary number $aperiod$ chosen from $[minP:maxP]$ interval before each alternance.}
\item {\textit{Local Optima GA (LocalOpt-GA)} (cf. Algo 3) : The alternance is realized if the population's best individual is a local optima. A boolean procedure {\bf Is\_Local\_Optima}({\bf Best\_Element}($pop$)) is used and returns true if $pop$ best individual is a local optima and false otherwise. A predifined subroutine {\bf Best\_Element}($pop$) is utilized to get the $pop$ best individual.}
\item {\textit{Homogeneous Population GA (HomogPop-GA)} (cf. Algo 4) : The alternance is realized if the population's standard deviation is less or equal to $\varepsilon$. A boolean procedure {\bf Is\_Homogeneous\_Population}($pop$, $\varepsilon$) is used and returns true if $pop$ standard deviation is less or equal to $\varepsilon$ and false otherwise.}
\item {\textit{Steady Generation GA (SteadyGen-GA)} (cf. Algo 5) : The alternance is realized if the population's best fitness value has not been changed for $steadyGen$ generations. A boolean procedure {\bf Is\_Steady\_Generation}($pop$, $steadyGen$) is used and returns true if $pop$ best fitness value has not been improved for $steadyGen$ generations and false otherwise.}
\end{enumerate}

\begin{algorithm}
\fontsize{8}{9}
\selectfont
\bf{Algorithm 1} \it{ Periodic-GA}
\hrule
\begin{algorithmic}
\STATE $period \leftarrow periodValue$
\STATE $coding \leftarrow starterCoding$
\STATE $pop \leftarrow$ {\bf Generate\_Initial\_Population}()
\REPEAT
	\REPEAT
		\STATE {\bf Run\_1\_SGA}($pop$, $coding$)
	\UNTIL{{\bf Is\_Period}($period$)}
	\STATE $coding \leftarrow$ {\bf Alternate\_Coding}($coding1$, $coding2$)
	\STATE {\bf Convert\_Population}($pop$, $coding$)
\UNTIL{{\bf Is\_MaxGen}($maxGen$)}
\end{algorithmic}
\end{algorithm}

\begin{algorithm}
\fontsize{8}{9}
\selectfont
\bf{Algorithm 2} \it{ Aperiodic-GA}
\hrule
\begin{algorithmic}
\STATE $coding \leftarrow starterCoding$
\STATE $pop \leftarrow$ {\bf Generate\_Initial\_Population}()
\REPEAT
	\STATE $aperiod \leftarrow Random[minP:maxP]$
	\REPEAT
		\STATE {\bf Run\_1\_SGA}($pop$, $coding$)
	\UNTIL{{\bf Is\_Period}($aperiod$)}
	\STATE $coding \leftarrow$ {\bf Alternate\_Coding}($coding1$, $coding2$)
	\STATE {\bf Convert\_Population}($pop$, $coding$)
\UNTIL{{\bf Is\_MaxGen}($maxGen$)}
\end{algorithmic}
\end{algorithm}

\begin{algorithm}
\fontsize{8}{9}
\selectfont
\bf{Algorithm 3} \it{ LocalOpt-GA}
\hrule
\begin{algorithmic}
\STATE $coding \leftarrow starterCoding$
\STATE $pop \leftarrow$ {\bf Generate\_Initial\_Population}()
\REPEAT
	\REPEAT
		\STATE {\bf Run\_1\_SGA}($pop$, $coding$)
	\UNTIL{{\bf Is\_Local\_Optima}({\bf Best\_Element}($pop$))}
	\STATE $coding \leftarrow$ {\bf Alternate\_Coding}($coding1$, $coding2$)
	\STATE {\bf Convert\_Population}($pop$, $coding$)
\UNTIL{{\bf Is\_MaxGen}($maxGen$)}
\end{algorithmic}
\end{algorithm}

\begin{algorithm}
\fontsize{8}{9}
\selectfont
\bf{Algorithm 4} \it{ HomogPop-GA}
\hrule
\begin{algorithmic}
\STATE $\varepsilon \leftarrow epsilon$
\STATE $coding \leftarrow starterCoding$
\STATE $pop \leftarrow$ {\bf Generate\_Initial\_Population}()
\REPEAT
	\REPEAT
		\STATE {\bf Run\_1\_SGA}($pop$, $coding$)
	\UNTIL{{\bf Is\_Homogeneous\_Population}($pop$, $\varepsilon$)}
	\STATE $coding \leftarrow$ {\bf Alternate\_Coding}($coding1$, $coding2$)
	\STATE {\bf Convert\_Population}($pop$, $coding$)
\UNTIL{{\bf Is\_MaxGen}($maxGen$)}
\end{algorithmic}
\end{algorithm}

\begin{algorithm}
\fontsize{8}{9}
\selectfont
\bf{Algorithm 5} \it{ SteadyGen-GA}
\hrule
\begin{algorithmic}
\STATE $steadyGen \leftarrow steadyGeneration$
\STATE $coding \leftarrow starterCoding$
\STATE $pop \leftarrow$ {\bf Generate\_Initial\_Population}()
\REPEAT
	\REPEAT
		\STATE {\bf Run\_1\_SGA}($pop$, $coding$)
	\UNTIL{{\bf Is\_Steady\_Generation}($pop$, $steadyGen$)}
	\STATE $coding \leftarrow$ {\bf Alternate\_Coding}($coding1$, $coding2$)
	\STATE {\bf Convert\_Population}($pop$, $coding$)
\UNTIL{{\bf Is\_MaxGen}($maxGen$)}
\end{algorithmic}
\end{algorithm}

\section{{\it SM-GA} Technique}
\label{secSMGAT}
\subsection{{\it SM-GA} Initiation}
Agents (units or sub-populations) are the entities, in literal meaning, that act or have the power of the authority to act on behalf of its designer. The basic and important features of the agents can be listed as autonomy, proactivity and collaboration, especially when we are designing agents to be used for representation utility. An autonomous agent works in a way that it can have self-activation mechanism and behaviour. Collaboration is a very important feature of an agent, which also makes an agent differ from an expert system. Collaboration gives the agent to communicate with other agents in the environment for either satisfying its goals or retrieving information in the environment. It is how that we got our first idea to the new formalism called {\it SM-GA}, a new technique planned on agents function and implemented in a resurgent encoding work engine in purpose to bring some order into the unsettled situation caused by the influence of representations on the performance of GAs.

\subsection{{\it SM-GA} Methodology of Work and Implementation}
\begin{figure}
\begin{center}
\epsfig{figure=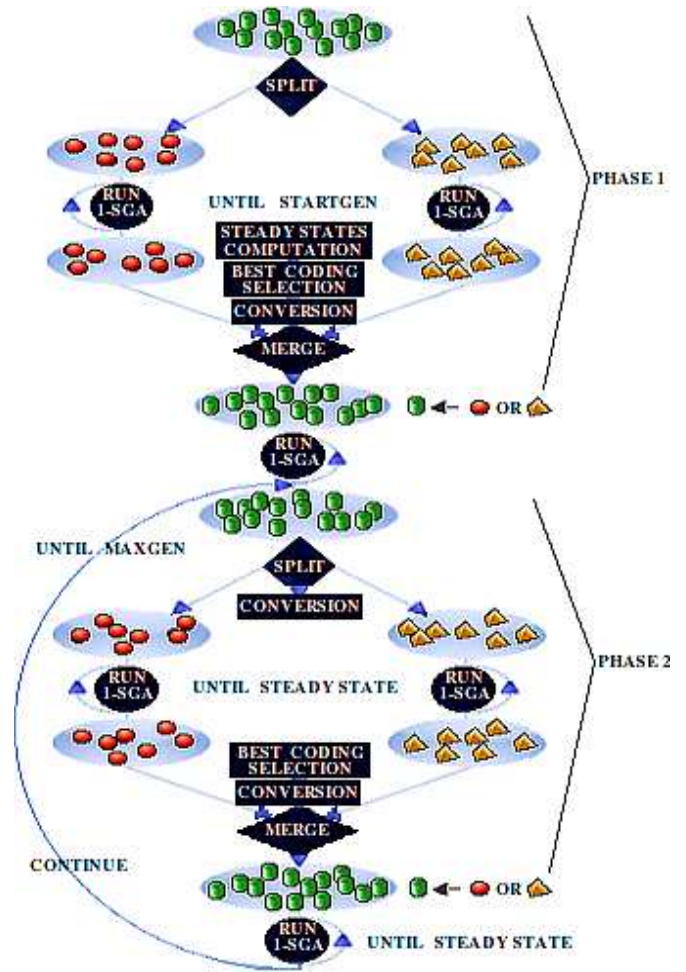,width=87mm,height=128.5mm}
\caption{{\it SM-GA} Schema.}
\label{figSMGA}
\end{center}
\end{figure}

{\it SM-GA} algorithm is based on the role of double-agents (dual coding). It includes two main phases and their functions can be resumed as follows : In first phase, this technique consists in generating randomly an initial population (first agent). Then, it belongs to splitting this basic population into two sub-populations (units) and getting each a distinct representation. Primarily, two synchronous SGAs are executed with these two units for a given number of generations ($startGen$). At this point, steady state (state of no improvement of best fitness value for a given number of generations) value is computed automatically for each coding. Steady state measurement for each representation is taken equal to the average of all steady states encountered during SGA operation in that representation for the $startGen$ generations. Then after the two units have achieved $startGen$ generations, it consists in merging all individuals in one population having a best coding representation. Best coding is selected relatively to the population that has the least average fitness. Next, an SGA is processed with the united population until meeting a steady state. After estimating regular values of steady states for each representation, second phase induces a re-splitting of the whole population into two sub-populations having each a different coding. Then, the two divided units are operated in parallel with two SGAs. In this manner, SGA will benefit from the two representations at the same time by the fact that this parallel genotype's codification describes proactivity appearing on two levels and evolution occurring on two scales simultaneously. Then after each generation, a test for steady state is necessary. If at least one of the two units encounters a corresponding steady state, then agents collaboration property will help to support and preserve landscaped the fitness productivity during the inquiry process. Thus, in a global manner, a merge of the two coexistent units into one unit having a best coding representation will be an appropriate and suitable issue in intention to gather and assemble all developed data. At this level, best individuals spread within the population and exchanges realized by crossover genetic operators and minor mutational changes in chromosomes make it possible for better structures to be generated. Next, an SGA will run with the integrated population until, at any rate, it deviates to a steady state probably caused by the existence of one or more local optima and which momently reveals its inability to make individuals evolve better. In that case, it consists in re-spliting the entire agent into two sub-agents, a simple idea induced by the fact of new-created agents will have respectively sufficient autonomy to auto-reshape and invert their unvarying pattern. By this way, possibly one of the two shrunk populations will have the opportunity to withdraw and surpass the local optima, a concept that will make it survive and retrieve its accurate direction to well discover the search space. Then, split-and-merge cycle continues until a given maximum number of generations $maxGen$ are attained (cf. Algo 6). The schema representing {\it SM-GA} whole process is shown in Fig \ref{figSMGA}. This algorithm parameter does not require fine tuning for each problem. Just, $startGen$ value must be large enough to be able to well estimate the steady states measurements for each coding. To optimize {\it SM-GA} algorithm, standard procedures were utilized. For given populations $pop$, $pop1$ and $pop2$, given representations $coding$, $coding1$ and $coding2$, and given numbers $steadyGen$ and $maxGen$, these procedures can be summarized as follows :
\begin{itemize}
\item {{\bf Split}($pop$, $pop1$, $pop2$) : takes $pop$ and divides it into two sub-populations $pop1$ and $pop2$.}
\item{{\bf Compute\_Steady\_State}($coding$, $startGen$) : estimates steady state value for $coding$ corresponding to the average of all steady states encountered while executing an SGA with $coding$ for $startGen$ generations.}
\item {{\bf Select\_Best\_Coding}($pop1$, $coding1$, $pop2$, $coding2$) : computes $pop1$ and $pop2$ fitness averages and returns the coding corresponding to the population that has the least average fitness.}
\item {{\bf Merge}($pop1$, $pop2$, $pop$) : takes $pop1$ and $pop2$ and blends them into $pop$.}
\end{itemize}

\begin{algorithm}
\fontsize{8}{9}
\selectfont
\bf{Algorithm 6} \it{ SM-GA}
\hrule
\begin{algorithmic}
\STATE $startGen \leftarrow startGeneration$
\STATE $pop \leftarrow$ {\bf Generate\_Initial\_Population}()
\STATE {\bf Split}($pop$, $pop1$, $pop2$)
\REPEAT
	\STATE {\bf Run\_1\_SGA}($pop1$, $coding1$)
	\STATE {\bf Run\_1\_SGA}($pop2$, $coding2$)
\UNTIL{{\bf Is\_Period}($startGen$)}
\STATE $steadyGen1 \leftarrow$ {\bf Compute\_Steady\_State}($coding1$, $startGen$)
\STATE $steadyGen2 \leftarrow$ {\bf Compute\_Steady\_State}($coding2$, $startGen$)
\STATE $bestCoding \leftarrow$ {\bf Select\_Best\_Coding}($pop1$, $coding1$, $pop2$, $coding2$)
\STATE {\bf Convert\_Population}($pop1$, $bestCoding$)
\STATE {\bf Convert\_Population}($pop2$, $bestCoding$)
\STATE {\bf Merge}($pop1$, $pop2$, $pop$)
\REPEAT
	\STATE {\bf Run\_1\_SGA}($pop$, $bestCoding$)
\UNTIL{{\bf Is\_Steady\_Generation}($pop$, $steadyGenOf(bestCoding)$)}
\REPEAT
	\STATE {\bf Split}($pop$, $pop1$, $pop2$)
	\STATE {\bf Convert\_Population}($pop1$, $coding1$)
	\STATE {\bf Convert\_Population}($pop2$, $coding2$)
	\REPEAT
		\STATE {\bf Run\_1\_SGA}($pop1$, $coding1$)
		\STATE {\bf Run\_1\_SGA}($pop2$, $coding2$)
	\UNTIL{{\bf Is\_Steady\_Generation}($pop1$, $steadyGen1$) {\bf or} {\bf Is\_Steady\_Generation}($pop2$, $steadyGen2$)}
	\STATE $bestCoding \leftarrow$ {\bf Select\_Best\_Coding}($pop1$, $coding1$, $pop2$, $coding2$)
	\STATE {\bf Convert\_Population}($pop1$, $bestCoding$)
	\STATE {\bf Convert\_Population}($pop2$, $bestCoding$)
	\STATE {\bf Merge}($pop1$, $pop2$, $pop$)
	\REPEAT
		\STATE {\bf Run\_1\_SGA}($pop$, $bestCoding$)
	\UNTIL{{\bf Is\_Steady\_Generation}($pop$, $steadyGenOf(bestCoding)$)}
\UNTIL{{\bf Is\_MaxGen}($maxGen$)}
\end{algorithmic}
\end{algorithm}

\section{Setup of Experiments}
\label{secSE}
\subsection{Test Problems}
\label{secTP}
Taking the most problematic and challenging test functions under consideration and given the nature of our study, we concluded to a total of five optimization functions. Table \ref{tabFunc} summarizes some of the unconstrained real-valued functions. All these routines are minimization problems and prove different degrees of complexity. Although, they were selected because of their ease of computation and widespread use, which should facilitate evaluation of the results.

The first test function Rosenbrock [F2] has been proposed by De Jong. It is unimodal (i.e. containing only one optimum) and is considered to be difficult because it has a very narrow ridge. The tip of the ridge is very sharp, and it runs around a parabola. Algorithms that are not able to discover good directions underperfom in this problem. Rosenbrock F2 has the global minimum at $(1, 1)$ \cite{r3}. The second function Schaffer [F6] has been conceived by Schaffer. It is an example of a multimodal function (i.e. containing many local optima, but only one global optimum) and is known to be a hard problem for GAs due to the number of local minima and the large search interval. Schaffer F6 has the global minimum at $(0, 0)$ and there are many nuisance local minima around it \cite{r3}. The third function Rastrigin [F7] is a typical model of a non-linear highly multimodal function. It is a fairly difficult problem for GAs due to the wide search space and the large number of local minima. It has a complexity of $\mathcal{O}(n\ln(n))$, where n is the number of function parameters. This function contains millions of local optima in the interval of consideration. Rastrigin F7 has the global minimum at $(0, ..., 0)$, i.e. in one corner of the search space \cite{r3}. The fourth function Griewangk [F8] also is a non-linear multimodal function. It has a complexity $\mathcal{O}(n\ln(n))$, where n is the number of function parameters. The terms of the summation produce a parabola, while the local optima are above parabola level. The dimensions of the search range increase on the basis of the product, which results in the decrease of the local minimums. The more we increase the search range, the ﬂatter the function. Generally speaking, this is a very difficult but good function for testing GAs performance mainly because the product creates sub-populations strongly codependent to parallel GAs models. Griewangk F8 has the global minimum at $(0, ..., 0)$ \cite{r3}. The fifth function Schwefel [F9] also is a non-linear multimodal function. It is somewhat easier than Rastrigin F7 and is characterized by a second-best minimum which is far away from the global optimum. In this function, V is the negative of the global minimum, which is added to the function so as to move the global minimum to zero, for convenience. The exact value of V depends on system precision; for our experiments V = 418.9829101. Schwefel F9 has the global minimum at $(420.9687, ..., 420.9687)$ \cite{r3}.

Most algorithms have difficulties to converge close to the minimum of such functions especially under high levels of dimensionality (i.e. in a black box form where the search algorithm should not necessarily assume independence of dimensions), because the probability of making progress decreases rapidly as the minimum is approached.

\begin{table}
\begin{center}
\caption{Objective Functions.}
\label{tabFunc}
\begin{tabular}{@{}l@{}l@{}l@{}}
\hline \\
{\bf Name} & \multicolumn{1}{c}{\bf Expression} & \hspace{0.6mm} {\bf Range} \hspace{5.2mm} {\bf Dimension} \\ \\
\hline \\
F2 & $f_2(x_i)=100 ( x_1^2 - x_2)^2 + (1 - x_1)^2$ & \hspace{2mm} $[-2.048:2.048]$ \hspace{2.05mm} $2$ \\ \\
F6 & $f_6(x_i)=0.5 + \frac{\sin^2(\sqrt{x^2 + y^2}) - 0.5}{(1 + 0.001 (x^2 + y^2))^2}$ & \hspace{2mm} $[-100:100]$ \hspace{6.62mm} $2$ \\ \\
F7 & $f_7(x_i)=200 + \sum\limits_{i=1}^{20} (x_i^2 - 10\cos(2{\pi}x_i))$ & \hspace{2mm} $[-5.12:5.12]$ \hspace{3.5mm} $20$ \\ \\
F8 & $f_8(x_i)=1 + \sum\limits_{i=1}^{10} (\frac{x_i^2}{4000}) - \prod\limits_{i=1}^{10} (\cos(\frac{x_i}{\sqrt{i}}))$ & \hspace{2mm} $[-600:600]$ \hspace{5.15mm} $10$ \\ \\
F9 & $f_9(x_i)=10V + \sum\limits_{i=1}^{10} (-x_i\sin(\sqrt{|x_i|}))$ & \hspace{2mm} $[-500:500]$ \hspace{5.15mm} $10$ \\ \\
\hline
\end{tabular}
\end{center}
\end{table}

\subsection{Parameter Settings}
In already defined proposals, an SGA was processed and it encapsulates the standard parameter values for any GA application which is based on binary strings representation. More specifically, the main common parameters are :
\begin{itemize}
\item {Pseudorandom generator : Uniform Generator.}
\item {Selection mechanism : Tournament Selection.}
\item {Crossover mechanism : 1-Point Crossover.}
\item {Mutation mechanism : Bit-Flip Mutation.}
\item {Replacement models : a) Generational Replacement. b) Elitism Replacement.}
\item {Algorithm ending criteria : the executions stop after maximum number of generations are reached.}
\end{itemize}
The set of remaining applied parameters are shown in Table \ref{tabParam} with : $maxGen$ for maximum number of generations before STOP, $popSize$ for population size, $vecSize$ for genotype size, $tSize$ for tournament selection size, $pCross$ for crossover rate, $1$-$PointRate$ for 1-point crossover rate, $pMut$ for mutation rate, $pMutPerBit$ for bit-flip mutation rate ($1 / vecSize$). Besides, the values of parameters necessary to new proposals are almost near for each function with a little difference evoked by the problem complexity. Values of these specific parameters were determined recurrently within fixed intervals lengths. In {\it Periodic-GA} and {\it Aperiodic-GA}, $period$ and $aperiod$ ([$minP:maxP$]) values changed within [$25:100$] interval with step of $5$. In {\it HomogPop-GA}, $\varepsilon$ value varied from $0.1$ to $5.0$ with step of $0.1$. In {\it SteadyGen-GA}, $steadyGen$ value changed within [$5:50$] interval with step of $5$. In {\it SM-GA}, $startGen$ value changed within [$100:500$] interval with step of $50$. Sufficient tests were performed to be able for attributing adequate values to each specific parameter. As has been discussed and after a large number of tests, we found that modifications of these parameters values within coherent fixed intervals lengths do not affect so much the final results of each proposal which accelerated a bit our plan of action. The best parameter settings between those tested are given in Table \ref{tabParam}.

\begin{table}
\begin{center}
\caption{Set of Used Parameters.}
\label{tabParam}
\begin{tabular}{|@{\hspace{1.25mm}}l@{\hspace{0.75mm}}|c|c|c|c|c|}
\hline
{\bf Parameters} & \multicolumn{5}{c|}{\bf Objective Functions} \\
\cline{2-6}
 & {\bf F2} & {\bf F6} & {\bf F7} & {\bf F8} & {\bf F9} \\
\hline
$maxGen$ & $3500$ & $3500$ & $3500$ & $3500$ & $3500$ \\
\hline
$popSize$ & $100$ & $100$ & $100$ & $100$ & $100$ \\
\hline
$vecSize$ & $40$ & $80$ & $200$ & $200$ & $150$ \\
\hline
$tSize$ & $2$ & $2$ & $4$ & $2$ & $2$ \\
\hline
$pCross$ & $0.6$ & $0.6$ & $1.0$ & $0.75$ & $0.6$ \\
\hline
$1$-$PointRate$ & $1.0$ & $1.0$ & $1.0$ & $1.0$ & $1.0$ \\
\hline
$pMut$ & $1.0$ & $1.0$ & $1.0$ & $1.0$ & $1.0$ \\
\hline
$pMutPerBit$ & $0.025$ & $0.0125$ & $0.0077$ & $0.0035$ & $0.006$ \\
\hline
$period$ & $50$ & $40$ & $25$ & $30$ & $10$ \\
\hline
[$minP:maxP$] & [$25$:$75$] & [$25$:$70$] & [$20$:$50$] & [$20$:$70$] & [$10$:$20$] \\
\hline
$\varepsilon$ & $5.0$ & $0.1$ & $5.0$ & $2.5$ & $1.0$ \\
\hline
$steadyGen$ & $35$ & $25$ & $5$ & $25$ & $5$ \\
\hline
$startGen$ & $250$ & $500$ & $100$ & $250$ & $250$ \\
\hline
\end{tabular}
\end{center}
\end{table}

\subsection{Testing Description and Numerical Observations}

\subsubsection{Real Numbers and Fitness Computation}
The real numbers are represented by binary bit strings of length $n*N$, where $n$ is the problem dimension and $N$ is the number of bits needed to represent each function parameter. $N$ is chosen in such as to have sufficient precision on the majority of real numbers included in the specific search space. In that case, the first $N$ bits represent the first parameter, the second $N$ bits represent the second parameter, and so forth. Given a function parameter $x$ represented by $N$ binary bits, if $x$ has an SC representation, then $x$ real value is computed by : $x = a + \frac{b - a}{2^N - 1} \sum_{i=0}^{N-1} x_i 2^i $ where $a$ and $b$ are respectively the minimum and maximum bounds of the search interval. If we write the Standard-binary-Coded value of a real $x$ as $s_{k-1}...s_1s_0$ and the Gray-Coded value as $g_{k-1}...g_1g_0$, then we have the relationships : $g_i = s_{i+1} \oplus s_i$ and $s_i = s_{i+1} \oplus g_i$ which allow conversion from one representation to the other (taking $s_k = 0$). In all cases and after real numbers computation, the fitness value was taken equal to the corresponding function value which was calculated according to the function expression given in Table \ref{tabFunc}.

\subsubsection{Experimental Results}
Testing new algorithms on objective functions, experimental results were reported in the limits to decide about the optimal proposal among all ones. Table \ref{tabRes} presents statistical results obtained over $200$ runs and at the last generation (gen $Nb$ $3500$). All problems are being minimized, this table shows generation number to optimum (GNTO) and succes rate (SR and SR2) results after $700000$ ($200$ $\times$ $3500$) executions for each proposal and each function, with the highest score in bold. GNTO value corresponds to the maximum number of generations needed to reach the optimum after entire process of all runs. SR value represents a percentage of the number of times the optimal solution is found after all executions. SR2 value represents a percentage of the number of times the optimal solution is found after all executions correspondingly to the minimum GNTO found for each function. For example, the minimum GNTO for function F9 was $2025$ recorded for {\it SM-GA} proposal; for {\it Periodic-GA}$_{SG}$ proposal, the GNTO was $3480$ and the corresponding SR was $100$, but if we wanted to note the SR measure of {\it Periodic-GA}$_{SG}$ found so far at generation numbered $2025$ we would detect a value of $79$ denoted SR2. In Table \ref{tabRes} : $_{SC}$ signifies an execution with SC, $_{GC}$ for an execution with GC, $_{SG}$ means that SC was the starter coding and $_{GS}$ when GC was the starter coding.

\begin{table*}
\fontsize{8}{9}
\selectfont
\begin{center}
\caption{Experimental Results.}
\label{tabRes}
\begin{tabular}{|@{\hspace{1mm}}l@{\hspace{0.3mm}}|@{}p{11.1mm}@{}|@{}p{9.4mm}@{}|@{}p{9.4mm}@{}||@{}p{11.1mm}@{}|@{}p{9.4mm}@{}|@{}p{9.4mm}@{}||@{}p{11.1mm}@{}|@{}p{9.4mm}@{}|@{}p{9.4mm}@{}||@{}p{11.1mm}@{}|@{}p{9.4mm}@{}|@{}p{9.4mm}@{}||@{}p{11.1mm}@{}|@{}p{9.4mm}@{}|@{}p{9.4mm}@{}|}
\hline
{\bf Proposal} & \multicolumn{3}{c||}{\bf F2} &  \multicolumn{3}{c||}{\bf F6} &  \multicolumn{3}{c||}{\bf F7} &  \multicolumn{3}{c||}{\bf F8} &  \multicolumn{3}{c|}{\bf F9} \\
\cline{2-16}
 & \multicolumn{1}{@{}c@{}|}{\bf GNTO} & \multicolumn{1}{@{}c@{}|}{\bf SR $\%$} & \multicolumn{1}{@{\hspace{0.25mm}}c@{\hspace{0.25mm}}||}{\bf SR2 $\%$} & \multicolumn{1}{@{}c@{}|}{\bf GNTO} & \multicolumn{1}{@{\hspace{0.85mm}}c@{\hspace{0.85mm}}|}{\bf SR $\%$} & \multicolumn{1}{@{\hspace{0.25mm}}c@{\hspace{0.25mm}}||}{\bf SR2 $\%$} & \multicolumn{1}{@{}c@{}|}{\bf GNTO} & \multicolumn{1}{@{}c@{}|}{\bf SR $\%$} & \multicolumn{1}{@{\hspace{0.25mm}}c@{\hspace{0.25mm}}||}{\bf SR2 $\%$} & \multicolumn{1}{@{}c@{}|}{\bf GNTO} & \multicolumn{1}{@{\hspace{0.85mm}}c@{\hspace{0.85mm}}|}{\bf SR $\%$} & \multicolumn{1}{@{\hspace{0.25mm}}c@{\hspace{0.25mm}}||}{\bf SR2 $\%$} & \multicolumn{1}{@{}c@{}|}{\bf GNTO} & \multicolumn{1}{@{}c@{}|}{\bf SR $\%$} & \multicolumn{1}{@{\hspace{0.25mm}}c@{\hspace{0.25mm}}|}{\bf SR2 $\%$} \\
\hline
{\it SGA}$_{SC}$ & \multicolumn{1}{c|}{$3500+$} & \multicolumn{1}{c|}{$32$} & \multicolumn{1}{c||}{$31$} & \multicolumn{1}{c|}{$3500+$} & \multicolumn{1}{c|}{$37$} & \multicolumn{1}{c||}{$37$} & \multicolumn{1}{c|}{$3500+$} & \multicolumn{1}{c|}{$1$} & \multicolumn{1}{c||}{$0$} & \multicolumn{1}{c|}{$3500+$} & \multicolumn{1}{c|}{$6$} & \multicolumn{1}{c||}{$6$} & \multicolumn{1}{c|}{$3500+$} & \multicolumn{1}{c|}{$0$} & \multicolumn{1}{c|}{$0$} \\
\hline
{\it SM-GA}$_{SC}$ & \multicolumn{1}{c|}{$3500+$} & \multicolumn{1}{c|}{$30$} & \multicolumn{1}{c||}{$30$} & \multicolumn{1}{c|}{$3500+$} & \multicolumn{1}{c|}{$48$} & \multicolumn{1}{c||}{$48$} & \multicolumn{1}{c|}{$3500+$} & \multicolumn{1}{c|}{$2$} & \multicolumn{1}{c||}{$2$} & \multicolumn{1}{c|}{$3500+$} & \multicolumn{1}{c|}{$11$} & \multicolumn{1}{c||}{$11$} & \multicolumn{1}{c|}{$3500+$} & \multicolumn{1}{c|}{$0$} & \multicolumn{1}{c|}{$0$} \\
\hline
{\it SGA}$_{GC}$ & \multicolumn{1}{c|}{$3500+$} & \multicolumn{1}{c|}{$99$} & \multicolumn{1}{c||}{$99$} & \multicolumn{1}{c|}{$3500+$} & \multicolumn{1}{c|}{$43$} & \multicolumn{1}{c||}{$43$} & \multicolumn{1}{c|}{$2957$} & \multicolumn{1}{c|}{$100$} & \multicolumn{1}{c||}{$99$} & \multicolumn{1}{c|}{$3500+$} & \multicolumn{1}{c|}{$3$} & \multicolumn{1}{c||}{$3$} & \multicolumn{1}{c|}{$2395$} & \multicolumn{1}{c|}{$100$} & \multicolumn{1}{c|}{$94$} \\
\hline
{\it SM-GA}$_{GC}$ & \multicolumn{1}{c|}{$3256$} & \multicolumn{1}{c|}{$100$} & \multicolumn{1}{c||}{$99$} & \multicolumn{1}{c|}{$3500+$} & \multicolumn{1}{c|}{$52$} & \multicolumn{1}{c||}{$52$} & \multicolumn{1}{c|}{$3362$} & \multicolumn{1}{c|}{$100$} & \multicolumn{1}{c||}{$98$} & \multicolumn{1}{c|}{$3500+$} & \multicolumn{1}{c|}{$8$} & \multicolumn{1}{c||}{$8$} & \multicolumn{1}{c|}{$2413$} & \multicolumn{1}{c|}{$100$} & \multicolumn{1}{c|}{$97$} \\
\hline
{\bf SM-GA} & \multicolumn{1}{c|}{\bf 3139} & \multicolumn{1}{c|}{\bf 100} & \multicolumn{1}{c||}{\bf 100} & \multicolumn{1}{c|}{\bf 3500+} & \multicolumn{1}{c|}{\bf 57} & \multicolumn{1}{c||}{\bf 57} & \multicolumn{1}{c|}{\bf 2940} & \multicolumn{1}{c|}{\bf 100} & \multicolumn{1}{c||}{\bf 100} & \multicolumn{1}{c|}{\bf 3500+} & \multicolumn{1}{c|}{\bf 17} & \multicolumn{1}{c||}{\bf 17} & \multicolumn{1}{c|}{\bf 2025} & \multicolumn{1}{c|}{\bf 100} & \multicolumn{1}{c|}{\bf 100} \\
\hline
\hline
{\bf Periodic-GA}$_{SG}$ & \multicolumn{1}{c|}{$3500+$} & \multicolumn{1}{c|}{$87$} & \multicolumn{1}{c||}{$86$} & \multicolumn{1}{c|}{$3500+$} & \multicolumn{1}{c|}{$51$} & \multicolumn{1}{c||}{$51$} & \multicolumn{1}{c|}{\bf 3191} & \multicolumn{1}{c|}{\bf 100} & \multicolumn{1}{c||}{\bf 98} & \multicolumn{1}{c|}{$3500+$} & \multicolumn{1}{c|}{$5$} & \multicolumn{1}{c||}{$5$} & \multicolumn{1}{c|}{$3480$} & \multicolumn{1}{c|}{$100$} & \multicolumn{1}{c|}{$79$} \\
\hline
{\bf Periodic-GA}$_{GS}$ & \multicolumn{1}{c|}{$3500+$} & \multicolumn{1}{c|}{$91$} & \multicolumn{1}{c||}{$91$} & \multicolumn{1}{c|}{$3500+$} & \multicolumn{1}{c|}{$46$} & \multicolumn{1}{c||}{$46$} & \multicolumn{1}{c|}{$3500+$} & \multicolumn{1}{c|}{$99$} & \multicolumn{1}{c||}{$80$} & \multicolumn{1}{c|}{$3500+$} & \multicolumn{1}{c|}{$4$} & \multicolumn{1}{c||}{$4$} & \multicolumn{1}{c|}{\bf 2279} & \multicolumn{1}{c|}{\bf 100} & \multicolumn{1}{c|}{\bf 92} \\
\hline
{\it Aperiodic-GA}$_{SG}$ & \multicolumn{1}{c|}{$3500+$} & \multicolumn{1}{c|}{$90$} & \multicolumn{1}{c||}{$90$} & \multicolumn{1}{c|}{$3500+$} & \multicolumn{1}{c|}{$51$} & \multicolumn{1}{c||}{$51$} & \multicolumn{1}{c|}{$3500+$} & \multicolumn{1}{c|}{$99$} & \multicolumn{1}{c||}{$93$} & \multicolumn{1}{c|}{$3500+$} & \multicolumn{1}{c|}{$3$} & \multicolumn{1}{c||}{$3$} & \multicolumn{1}{c|}{$3009$} & \multicolumn{1}{c|}{$100$} & \multicolumn{1}{c|}{$89$} \\
\hline
{\bf Aperiodic-GA}$_{GS}$ & \multicolumn{1}{c|}{$3500+$} & \multicolumn{1}{c|}{$93$} & \multicolumn{1}{c||}{$92$} & \multicolumn{1}{c|}{\bf 3500+} & \multicolumn{1}{c|}{\bf 52} & \multicolumn{1}{c||}{\bf 52} & \multicolumn{1}{c|}{$3230$} & \multicolumn{1}{c|}{$100$} & \multicolumn{1}{c||}{$93$} & \multicolumn{1}{c|}{$3500+$} & \multicolumn{1}{c|}{$3$} & \multicolumn{1}{c||}{$3$} & \multicolumn{1}{c|}{$2818$} & \multicolumn{1}{c|}{$100$} & \multicolumn{1}{c|}{$90$} \\
\hline
{\it LocalOpt-GA}$_{SG}$ & \multicolumn{1}{c|}{$3500+$} & \multicolumn{1}{c|}{$78$} & \multicolumn{1}{c||}{$76$} & \multicolumn{1}{c|}{$3500+$} & \multicolumn{1}{c|}{$49$} & \multicolumn{1}{c||}{$49$} & \multicolumn{1}{c|}{$3500+$} & \multicolumn{1}{c|}{$96$} & \multicolumn{1}{c||}{$88$} & \multicolumn{1}{c|}{$3500+$} & \multicolumn{1}{c|}{$6$} & \multicolumn{1}{c||}{$6$} & \multicolumn{1}{c|}{$2480$} & \multicolumn{1}{c|}{$100$} & \multicolumn{1}{c|}{$93$} \\
\hline
{\it LocalOpt-GA$_{GS}$} & \multicolumn{1}{c|}{$3500+$} & \multicolumn{1}{c|}{$81$} & \multicolumn{1}{c||}{$80$} & \multicolumn{1}{c|}{$3500+$} & \multicolumn{1}{c|}{$48$} & \multicolumn{1}{c||}{$48$} & \multicolumn{1}{c|}{$3491$} & \multicolumn{1}{c|}{$100$} & \multicolumn{1}{c||}{$95$} & \multicolumn{1}{c|}{$3500+$} & \multicolumn{1}{c|}{$5$} & \multicolumn{1}{c||}{$5$} & \multicolumn{1}{c|}{$2480$} & \multicolumn{1}{c|}{$100$} & \multicolumn{1}{c|}{$94$} \\
\hline
{\it HomogPop-GA}$_{SG}$ & \multicolumn{1}{c|}{$3500+$} & \multicolumn{1}{c|}{$32$} & \multicolumn{1}{c||}{$31$} & \multicolumn{1}{c|}{$3500+$} & \multicolumn{1}{c|}{$41$} & \multicolumn{1}{c||}{$41$} & \multicolumn{1}{c|}{$3500+$} & \multicolumn{1}{c|}{$1$} & \multicolumn{1}{c||}{$0$} & \multicolumn{1}{c|}{$3500+$} & \multicolumn{1}{c|}{$2$} & \multicolumn{1}{c||}{$2$} & \multicolumn{1}{c|}{$3500+$} & \multicolumn{1}{c|}{$0$} & \multicolumn{1}{c|}{$0$} \\
\hline
{\bf HomogPop-GA}$_{GS}$ & \multicolumn{1}{c|}{\bf 3500+} & \multicolumn{1}{c|}{\bf 99} & \multicolumn{1}{c||}{\bf 99} & \multicolumn{1}{c|}{$3500+$} & \multicolumn{1}{c|}{$38$} & \multicolumn{1}{c||}{$38$} & \multicolumn{1}{c|}{$3001$} & \multicolumn{1}{c|}{$100$} & \multicolumn{1}{c||}{$95$} & \multicolumn{1}{c|}{$3500+$} & \multicolumn{1}{c|}{$3$} & \multicolumn{1}{c||}{$3$} & \multicolumn{1}{c|}{$2395$} & \multicolumn{1}{c|}{$100$} & \multicolumn{1}{c|}{$94$} \\
\hline
{\bf SteadyGen-GA}$_{SG}$ & \multicolumn{1}{c|}{$3500+$} & \multicolumn{1}{c|}{$88$} & \multicolumn{1}{c||}{$87$} & \multicolumn{1}{c|}{$3500+$} & \multicolumn{1}{c|}{$46$} & \multicolumn{1}{c||}{$46$} & \multicolumn{1}{c|}{$3381$} & \multicolumn{1}{c|}{$100$} & \multicolumn{1}{c||}{$95$} & \multicolumn{1}{c|}{\bf 3500+} & \multicolumn{1}{c|}{\bf 7} & \multicolumn{1}{c||}{\bf 7} & \multicolumn{1}{c|}{$2453$} & \multicolumn{1}{c|}{$100$} & \multicolumn{1}{c|}{$86$} \\
\hline
{\it SteadyGen-GA}$_{GS}$ & \multicolumn{1}{c|}{$3500+$} & \multicolumn{1}{c|}{$89$} & \multicolumn{1}{c||}{$88$} & \multicolumn{1}{c|}{$3500+$} & \multicolumn{1}{c|}{$51$} & \multicolumn{1}{c||}{$51$} & \multicolumn{1}{c|}{$3254$} & \multicolumn{1}{c|}{$100$} & \multicolumn{1}{c||}{$98$} & \multicolumn{1}{c|}{$3500+$} & \multicolumn{1}{c|}{$4$} & \multicolumn{1}{c||}{$4$} & \multicolumn{1}{c|}{$2894$} & \multicolumn{1}{c|}{$100$} & \multicolumn{1}{c|}{$82$} \\
\hline
\end{tabular}
\end{center}
\end{table*}

\subsubsection{Student's t-test}
Generally, the Student's t-test serves for comparing the means of two experiences and assesses whether they are statistically different from each other. As well in our experiments, the t-test was used to compare, across all runs, success rate (SR2) and mean best fitness (MBF) results between different proposals so it will help to judge the difference between their averages relative to the spread or variability of their scores. Regarding Table \ref{tabRes} which distinctly shows the performances of {\it SM-GA} towards other proposals, t-test results were studied in comparison between {\it SM-GA} and other algorithms. Computed results are displayed in Table \ref{tabttRes}.

\begin{table}
\begin{center}
\caption{t-test Results : Comparison between {\it SM-GA} and other Algorithms.}
\label{tabttRes}
\begin{tabular}{|@{\hspace{0.7mm}}l@{\hspace{0.1mm}}|@{}p{6.1mm}@{}|@{}p{6.1mm}@{}||@{}p{6.1mm}@{}|@{}p{6.1mm}@{}||@{}p{6.1mm}@{}|@{}p{6.1mm}@{}||@{}p{6.1mm}@{}|@{}p{6.1mm}@{}||@{}p{6.1mm}@{}|@{}p{6.1mm}@{}|}
\hline
{\bf SM-GA} & \multicolumn{2}{c||}{\bf F2} & \multicolumn{2}{c||}{\bf F6} & \multicolumn{2}{c||}{\bf F7} & \multicolumn{2}{c||}{\bf F8} & \multicolumn{2}{c|}{\bf F9} \\
\cline{2-11}
{\bf Compared to} & \hspace{0.6mm}{\bf SR2} &  {\bf MBF} & \hspace{0.6mm}{\bf SR2} &  {\bf MBF} & \hspace{0.6mm}{\bf SR2} &  {\bf MBF} & \hspace{0.6mm}{\bf SR2} &  {\bf MBF} & \hspace{0.6mm}{\bf SR2} &  {\bf MBF} \\
\hline
{\it SGA}$_{SC}$ & \hspace{1.5mm}$21$ & \hspace{1.4mm}$10$ & \hspace{1.1mm}$4.1$ & \hspace{1.1mm}$5.1$ & \hspace{1.1mm}$inf$ & \hspace{1.4mm}$25$ & \hspace{1.1mm}$3.6$ & \hspace{1.4mm}$11$ & \hspace{1.1mm}$inf$ & \hspace{1.4mm}$47$ \\
\hline
{\it SM-GA}$_{SC}$ & \hspace{1.5mm}$21$ & \hspace{1.1mm}$8.7$ & \hspace{1.1mm}$1.9$ & \hspace{1.1mm}$2.6$ & \hspace{1.4mm}$98$ & \hspace{1.4mm}$24$ & \hspace{1.1mm}$1.8$ & \hspace{1.1mm}$0.5$ & \hspace{1.1mm}$inf$ & \hspace{1.4mm}$42$ \\
\hline
{\it SGA}$_{GC}$ & \hspace{1.1mm}$1.5$ & \hspace{1.1mm}$1.1$ & \hspace{1.1mm}$2.9$ & \hspace{1.1mm}$4.5$ & \hspace{1.1mm}$1.5$ & \hspace{1.1mm}$1.1$ & \hspace{1.1mm}$4.8$ & \hspace{1.1mm}$9.3$ & \hspace{1.1mm}$3.6$ & \hspace{1.1mm}$2.5$ \\
\hline
{\it SM-GA}$_{GC}$ & \hspace{1.1mm}$1.5$ & \hspace{1.1mm}$1.1$ & \hspace{1.1mm}$1.1$ & \hspace{1.1mm}$3.1$ & \hspace{1.1mm}$2.1$ & \hspace{1.1mm}$1.6$ & \hspace{1.1mm}$2.8$ & \hspace{1.1mm}$3.1$ & \hspace{1.1mm}$2.5$ & \hspace{1.1mm}$1.5$ \\
\hline
\hline
{\it Periodic-GA}$_{SG}$ & \hspace{1.1mm}$5.8$ & \hspace{1.1mm}$5.1$ & \hspace{1.1mm}$1.3$ & \hspace{1.1mm}$2.1$ & \hspace{1.1mm}$2.1$ & \hspace{1.1mm}$0.3$ & \hspace{1.1mm}$3.9$ & \hspace{1.1mm}$3.9$ & \hspace{1.1mm}$7.3$ & \hspace{1.1mm}$3.2$ \\
\hline
{\it Periodic-GA}$_{GS}$ & \hspace{1.1mm}$4.5$ & \hspace{1.1mm}$4.3$ & \hspace{1.1mm}$2.3$ & \hspace{1.1mm}$2.7$ & \hspace{1.1mm}$7.1$ & \hspace{1.1mm}$6.8$ & \hspace{1.1mm}$4.4$ & \hspace{1.1mm}$4.4$ & \hspace{1.1mm}$4.1$ & \hspace{1.1mm}$3.9$ \\
\hline
{\it Aperiodic-GA}$_{SG}$ & \hspace{1.1mm}$4.8$ & \hspace{1.1mm}$4.2$ & \hspace{1.1mm}$1.3$ & \hspace{1.1mm}$2.1$ & \hspace{1.1mm}$3.9$ & \hspace{1.1mm}$3.2$ & \hspace{1.1mm}$4.8$ & \hspace{1.1mm}$5.1$ & \hspace{1.1mm}$4.9$ & \hspace{1.1mm}$2.2$ \\
\hline
{\it Aperiodic-GA}$_{GS}$ & \hspace{1.1mm}$4.2$ & \hspace{1.1mm}$1.7$ & \hspace{1.1mm}$1.1$ & \hspace{1.1mm}$1.7$ & \hspace{1.1mm}$3.9$ & \hspace{1.1mm}$2.7$ & \hspace{1.1mm}$4.8$ & \hspace{1.1mm}$3.3$ & \hspace{1.1mm}$4.8$ & \hspace{1.1mm}$1.9$ \\
\hline
{\it LocalOpt-GA}$_{SG}$ & \hspace{1.1mm}$7.9$ & \hspace{1.1mm}$4.6$ & \hspace{1.1mm}$1.7$ & \hspace{1.1mm}$2.7$ & \hspace{1.1mm}$5.3$ & \hspace{1.1mm}$4.8$ & \hspace{1.1mm}$3.6$ & \hspace{1.1mm}$3.3$ & \hspace{1.1mm}$3.9$ & \hspace{1.1mm}$1.5$ \\
\hline
{\it LocalOpt-GA}$_{GS}$ & \hspace{1.1mm}$7.1$ & \hspace{1.1mm}$5.1$ & \hspace{1.1mm}$1.9$ & \hspace{1.1mm}$2.7$ & \hspace{1.1mm}$3.3$ & \hspace{1.1mm}$2.3$ & \hspace{1.1mm}$3.9$ & \hspace{1.1mm}$5.8$ & \hspace{1.1mm}$3.6$ & \hspace{1.1mm}$2.6$ \\
\hline
{\it HomogPop-GA}$_{SG}$ & \hspace{1.4mm}$21$ & \hspace{1.4mm}$10$ & \hspace{1.1mm}$3.3$ & \hspace{1.1mm}$3.4$ & \hspace{1.1mm}$inf$ & \hspace{1.4mm}$25$ & \hspace{1.1mm}$5.3$ & \hspace{1.1mm}$6.1$ & \hspace{1.1mm}$inf$ & \hspace{1.4mm}$47$ \\
\hline
{\it HomogPop-GA}$_{GS}$ & \hspace{1.1mm}$1.5$ & \hspace{1.1mm}$1.1$ & \hspace{1.1mm}$3.9$ & \hspace{1.1mm}$4.7$ & \hspace{1.1mm}$3.3$ & \hspace{1.1mm}$1.1$ & \hspace{1.1mm}$4.8$ & \hspace{1.1mm}$5.5$ & \hspace{1.1mm}$3.6$ & \hspace{1.1mm}$2.5$ \\
\hline
{\it SteadyGen-GA}$_{SG}$ & \hspace{1.1mm}$5.5$ & \hspace{1.1mm}$4.9$ & \hspace{1.1mm}$2.3$ & \hspace{1.1mm}$3.1$ & \hspace{1.1mm}$3.3$ & \hspace{1.1mm}$2.5$ & \hspace{1.1mm}$3.2$ & \hspace{1.1mm}$4.1$ & \hspace{1.1mm}$5.8$ & \hspace{1.1mm}$2.4$ \\
\hline
{\it SteadyGen-GA}$_{GS}$ & \hspace{1.1mm}$5.3$ & \hspace{1.1mm}$4.7$ & \hspace{1.1mm}$1.3$ & \hspace{1.1mm}$1.5$ & \hspace{1.1mm}$2.1$ & \hspace{1.1mm}$1.4$ & \hspace{1.1mm}$4.4$ & \hspace{1.1mm}$6.1$ & \hspace{1.1mm}$6.7$ & \hspace{1.1mm}$2.9$ \\
\hline
\end{tabular}
\end{center}
\end{table}

\begin{figure}
\begin{tabular}{@{}c@{}c@{}}
\epsfig{figure=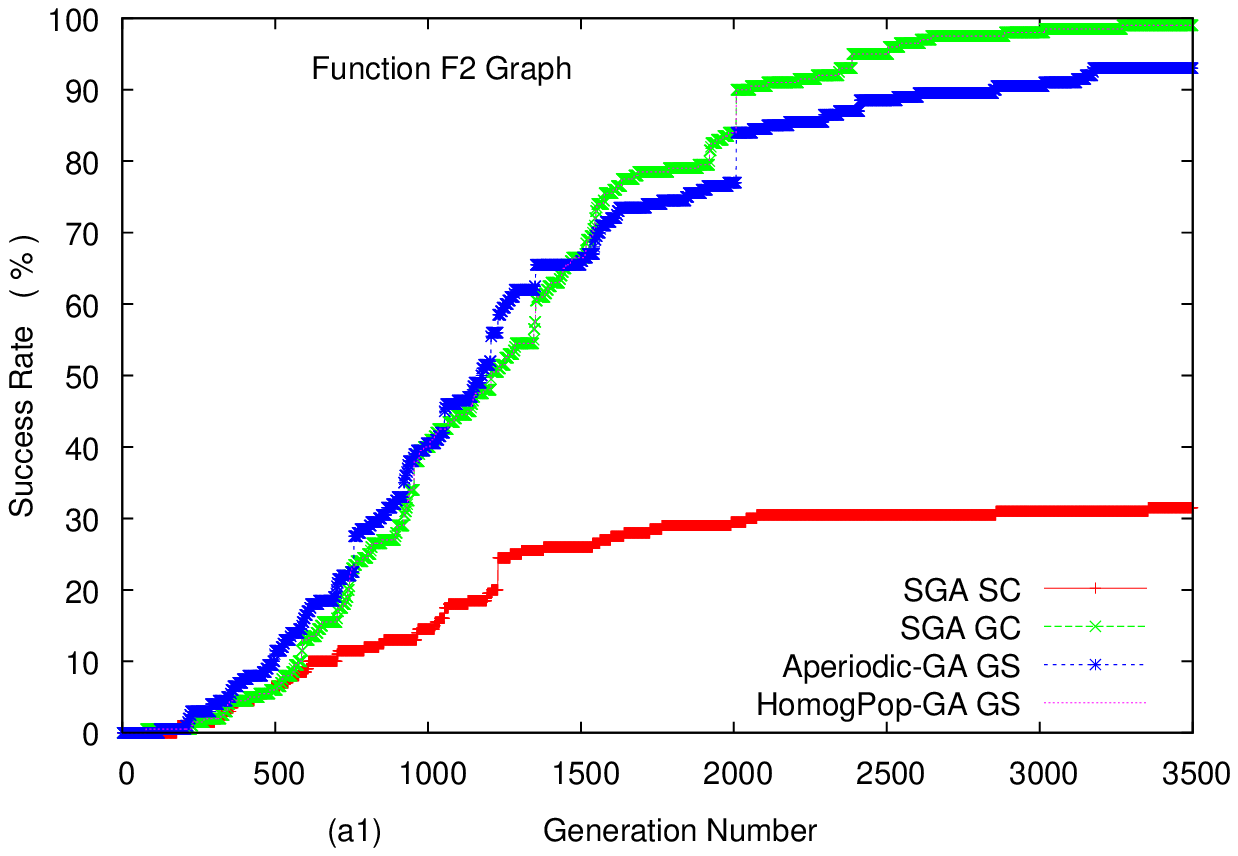, width=44.64mm, height=34.12mm} & \epsfig{figure=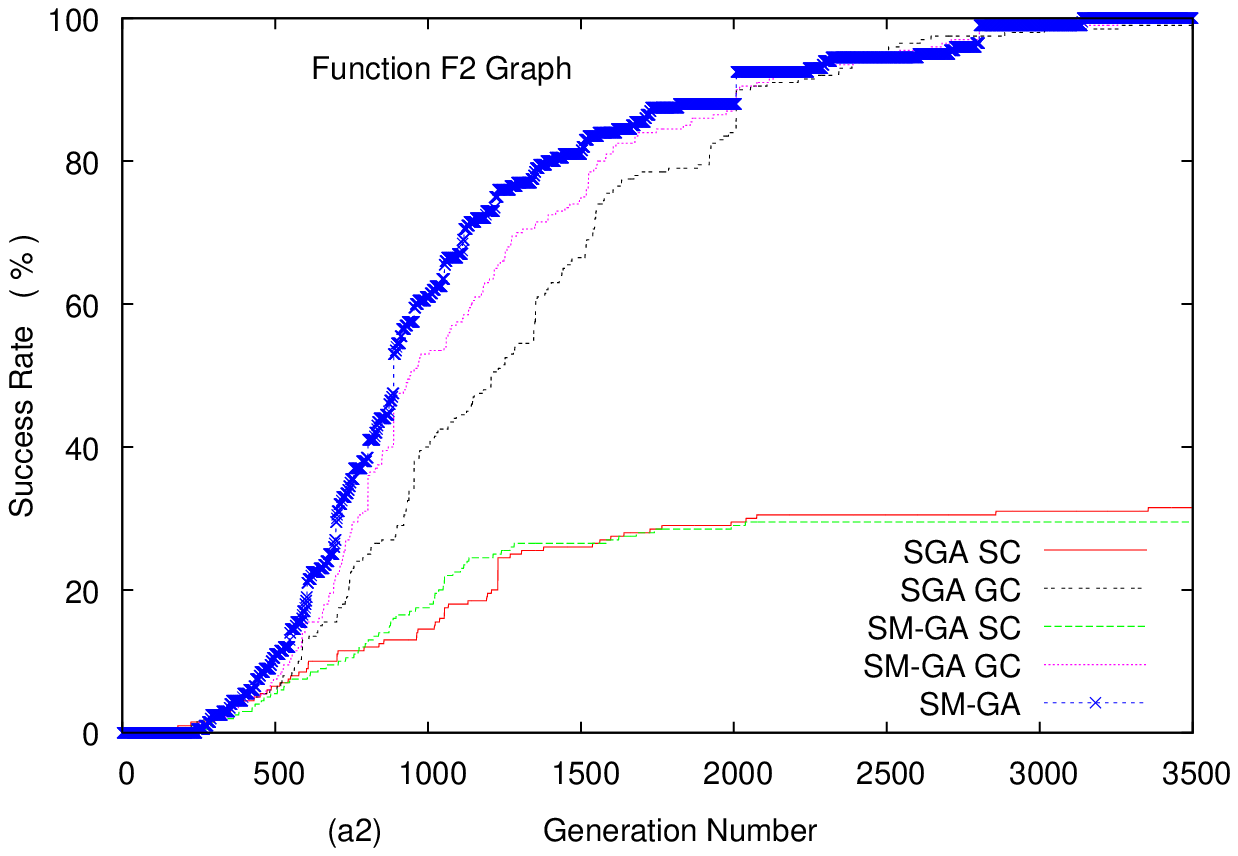, width=44.64mm, height=34.12mm} \\
\epsfig{figure=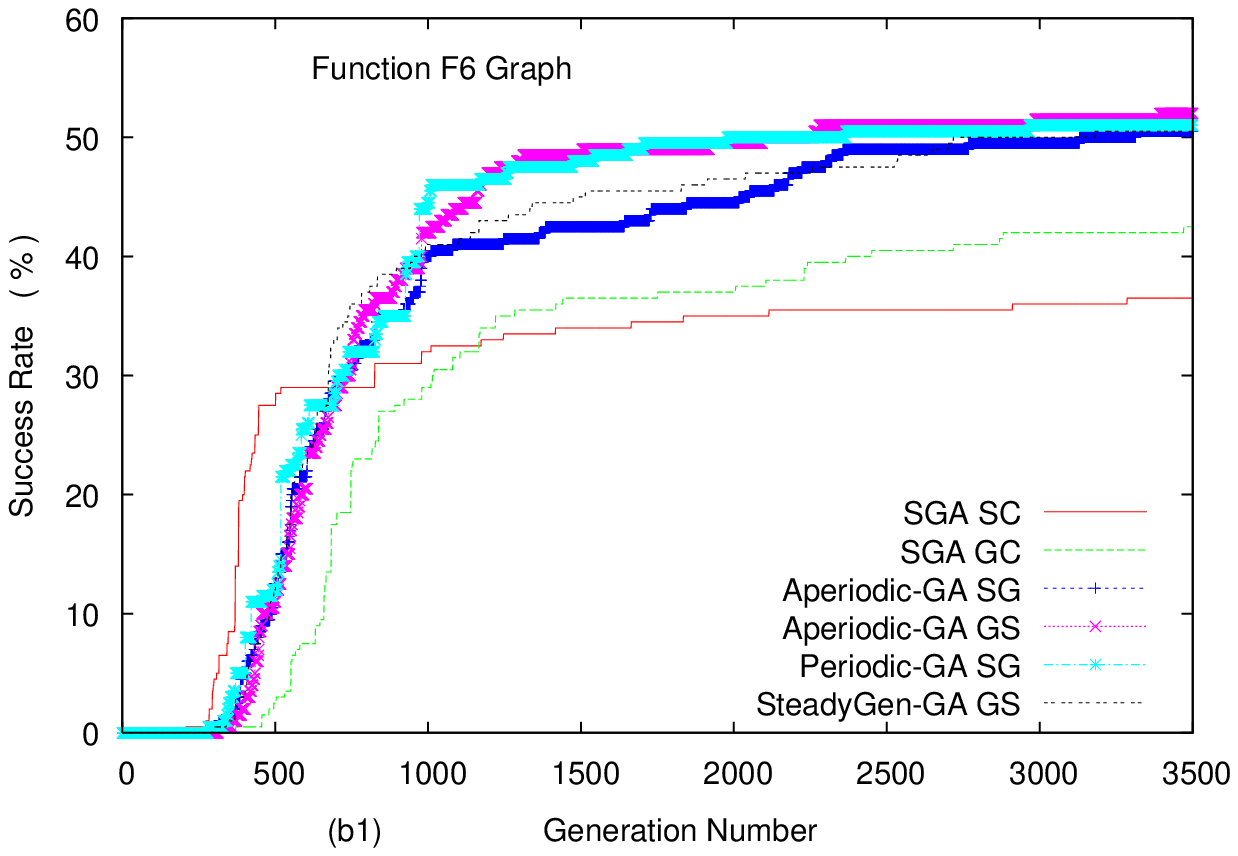, width=44.64mm, height=34.12mm} & \epsfig{figure=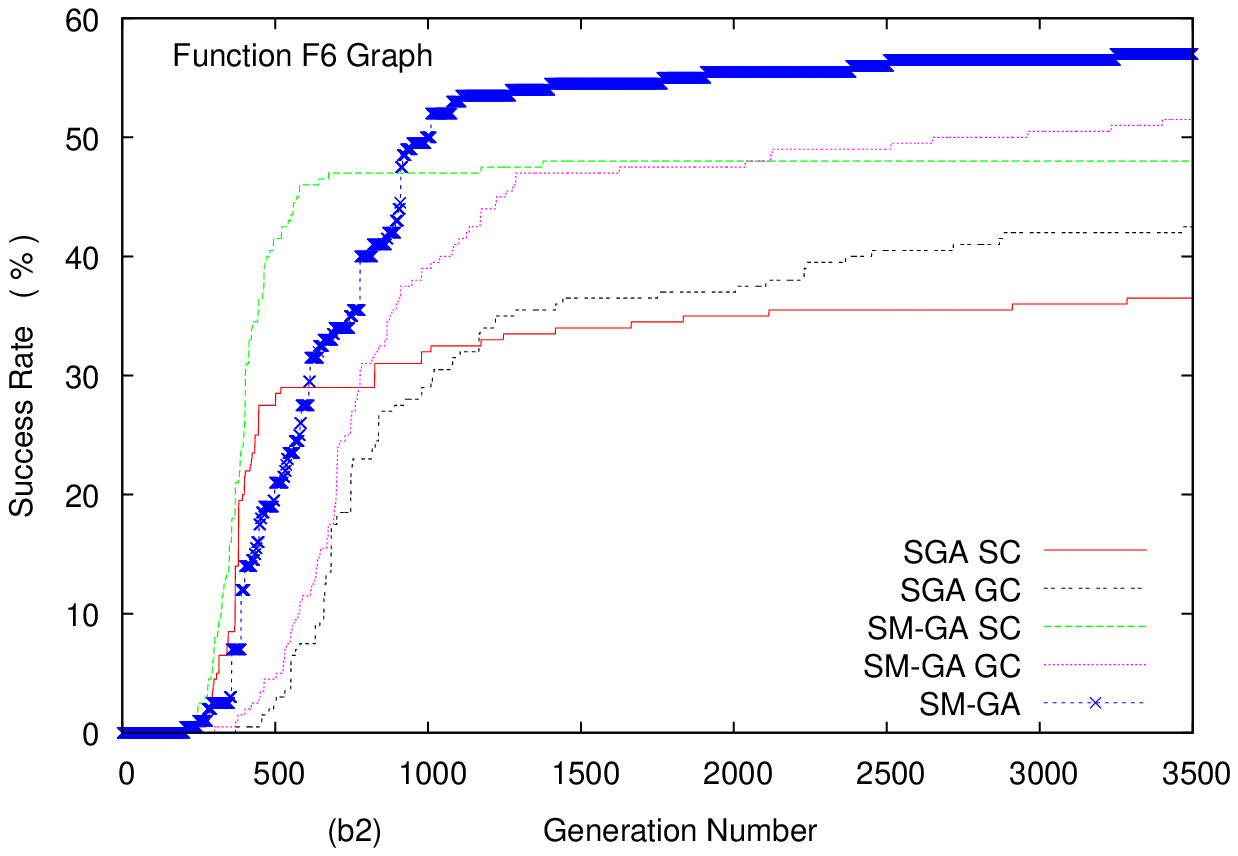, width=44.64mm, height=34.12mm} \\
\epsfig{figure=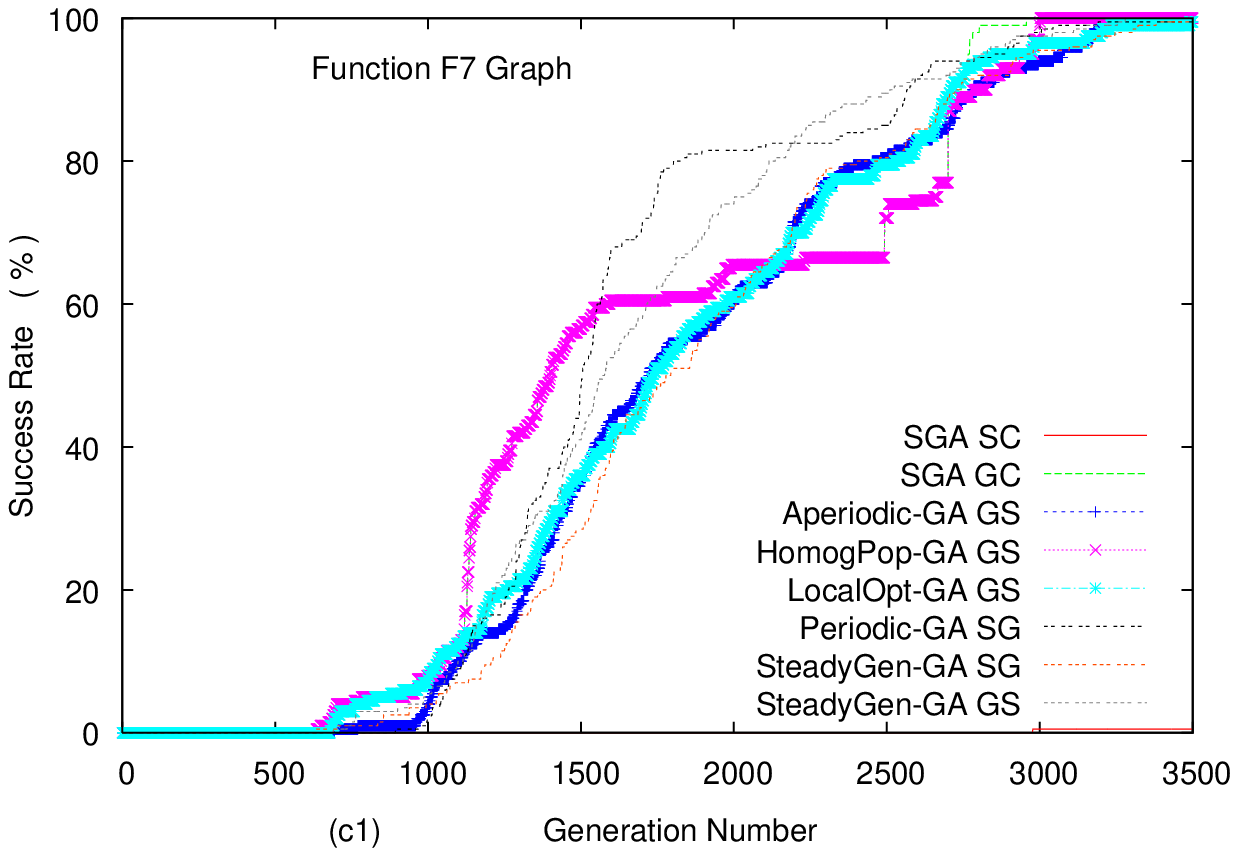, width=44.64mm, height=34.12mm} & \epsfig{figure=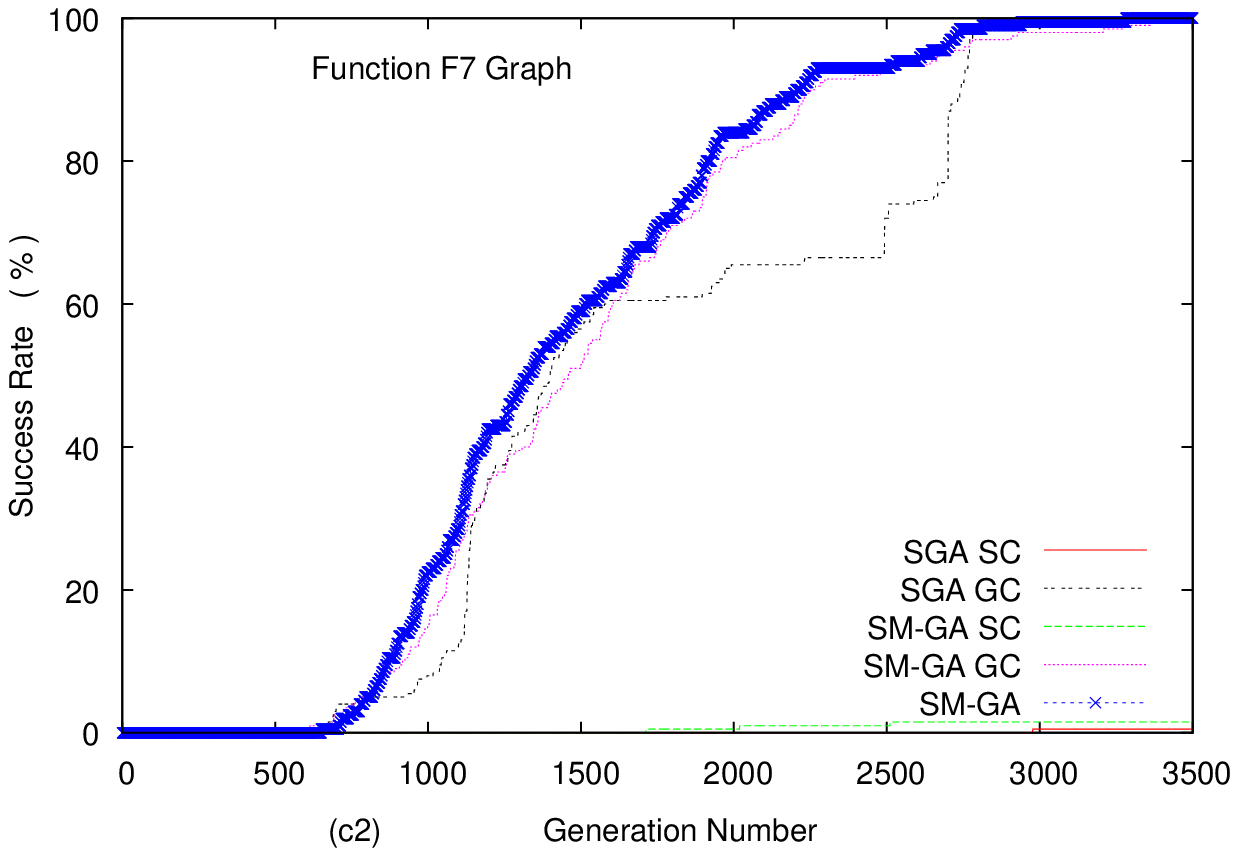, width=44.64mm, height=34.12mm} \\
\epsfig{figure=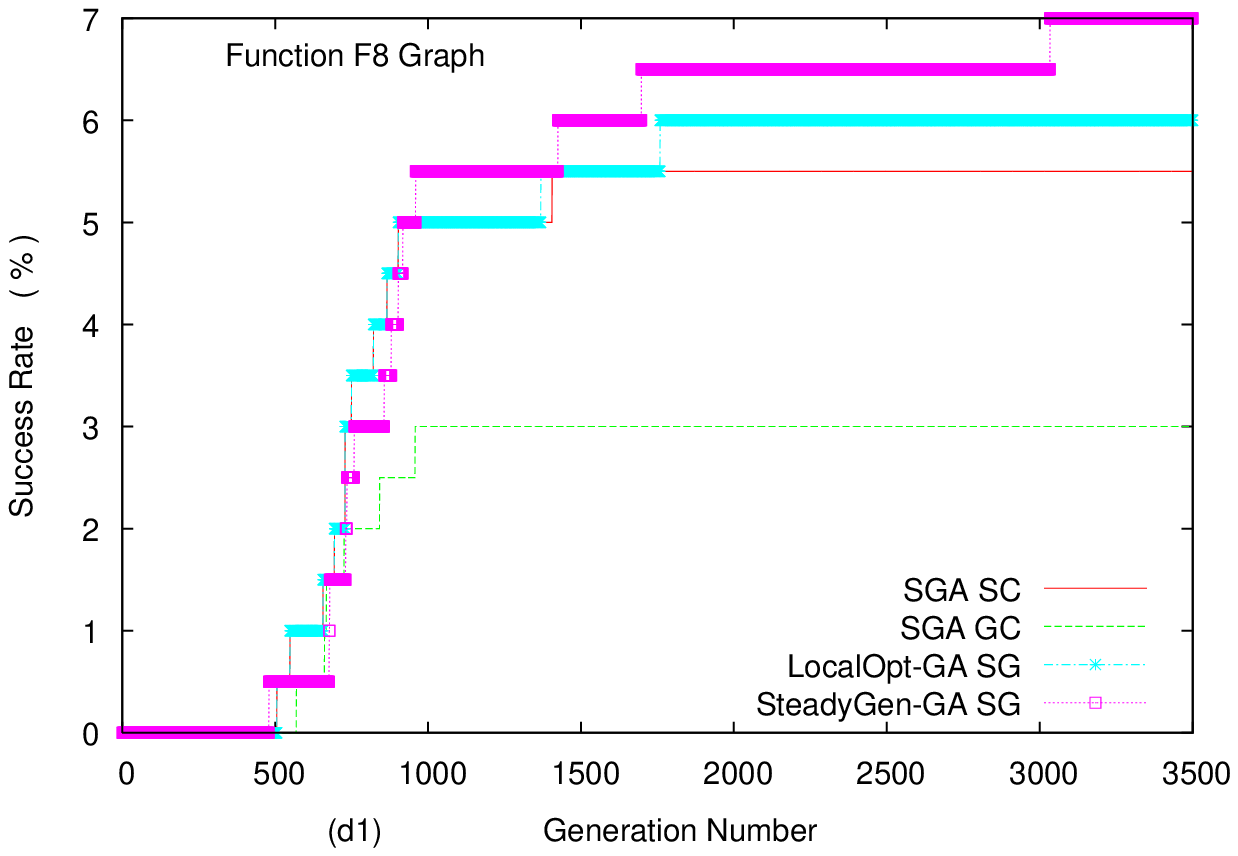, width=44.64mm, height=34.12mm} & \epsfig{figure=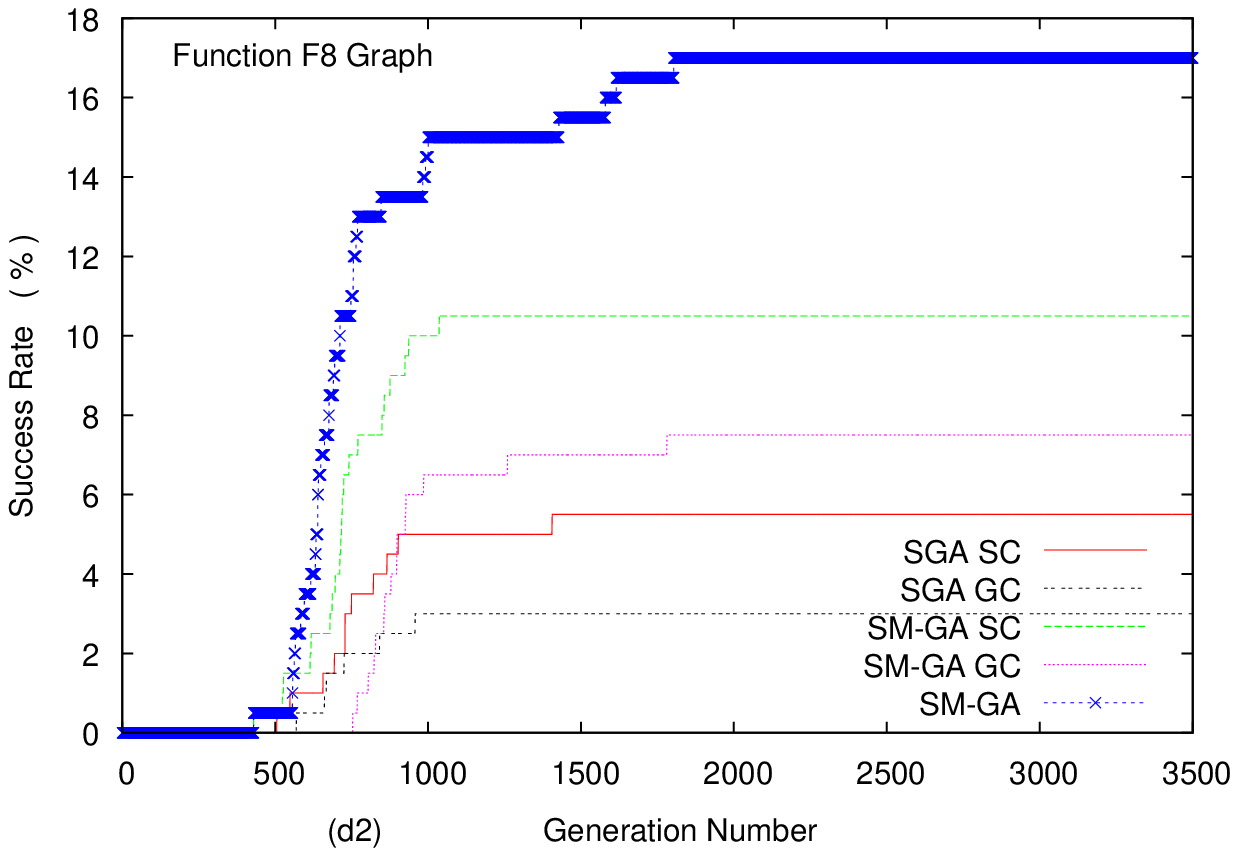, width=44.64mm, height=34.12mm} \\
\epsfig{figure=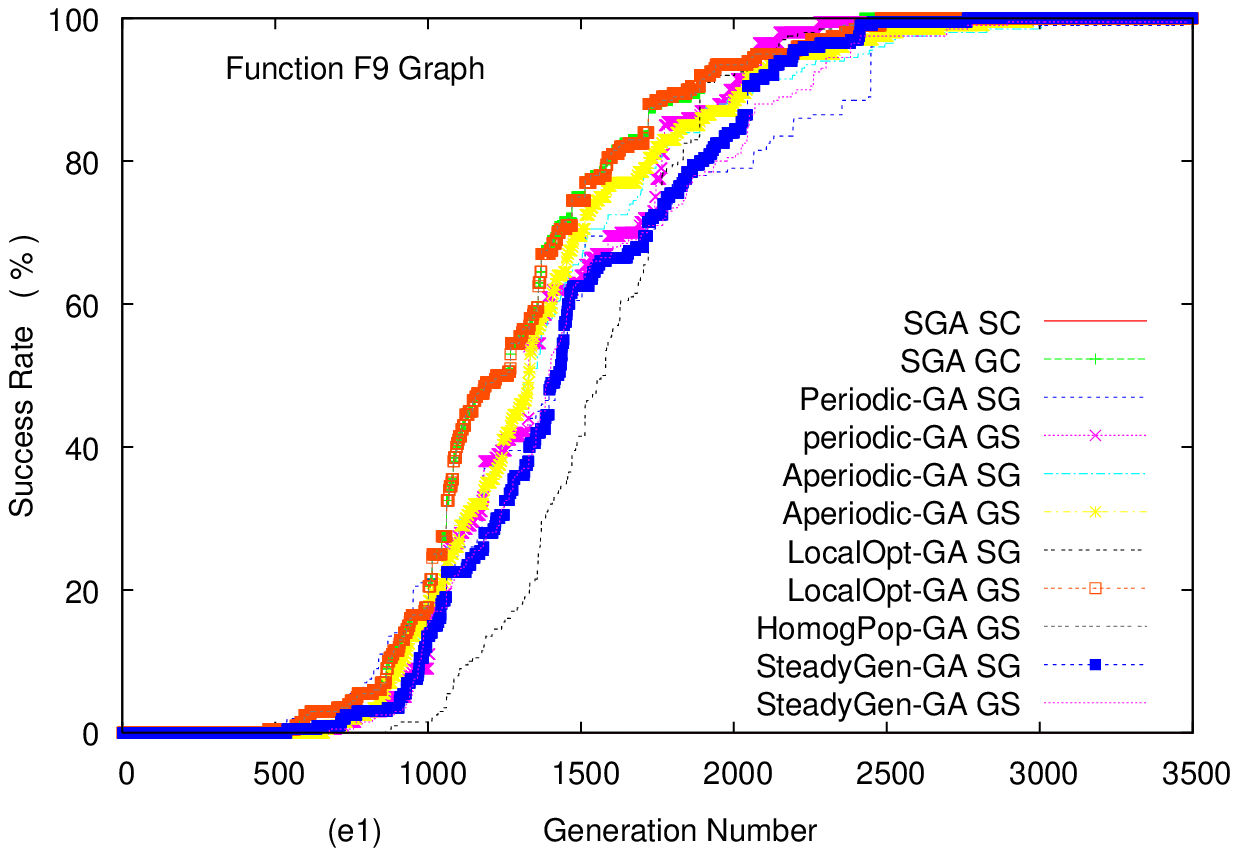, width=44.64mm, height=34.12mm} & \epsfig{figure=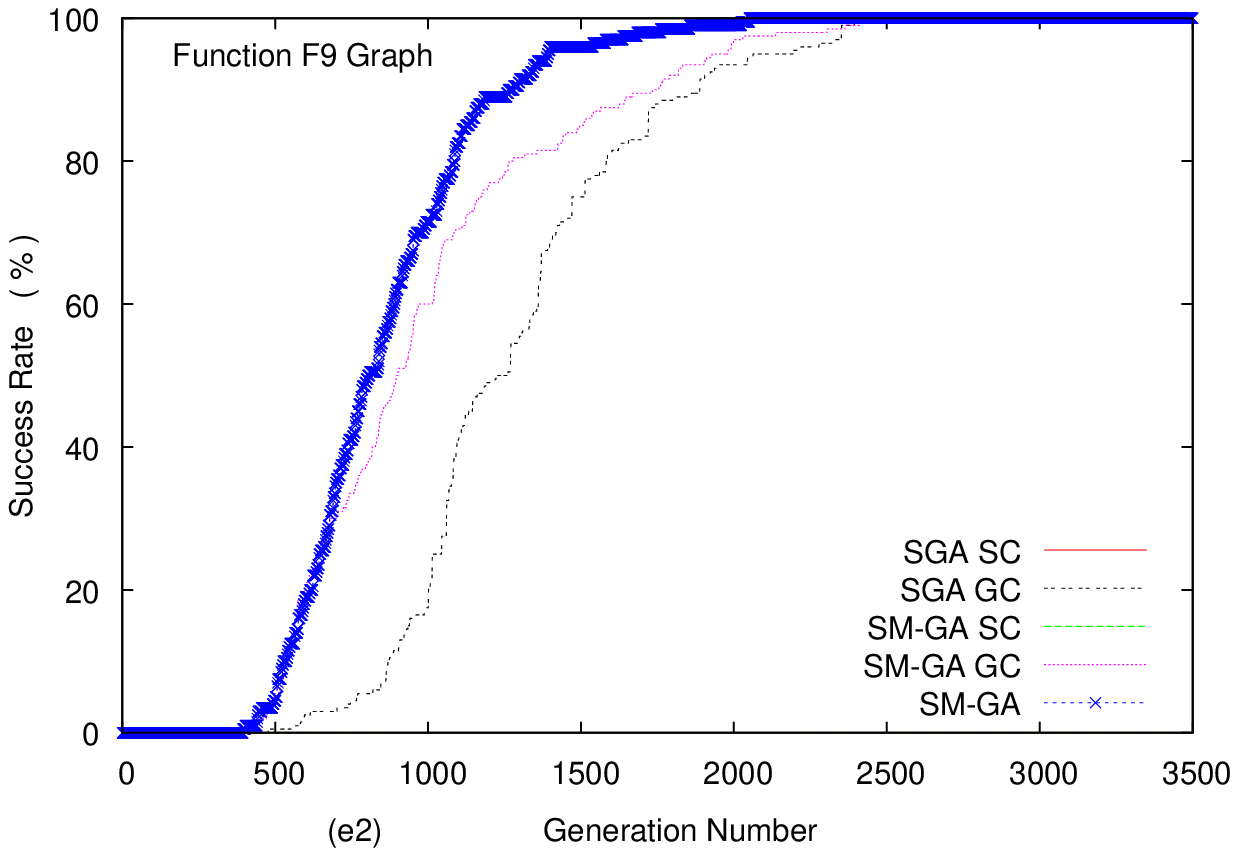, width=44.64mm, height=34.12mm}
\end{tabular}
\caption{Success Rate Evolution over Generations : Comparison between Different Proposals.}
\label{figSR}
\end{figure}

\section{General Discussions and Conclusion}
\label{secGDC}
Genetic algorithms, as has been discussed, provide a very good conceptional framework for optimization inspired by nature, but theoretical questions and algorithmic considerations deliberated in this work suggest that an SGA with static single coding sometimes fails to converge to the desired solution in a defined number of generations, a state called GA {\it deception} in optimization task, by the fact that selecting a representation that conflicts and opposes to a problem's fitness function can make that problem much hard and difficult for a GA to solve. In this paper, we tried to make SC and GC interacts each with other to transform the binary parameter representation for the problem to avoid compromising the difficulty of the problem because both SC and GC produce all possible representations and both have quite a lot of advantages. Yet, we started by formulating Serial Dual Coding strategies in a dynamic manner to study the fundamental interaction while alternating between two representations. Likewise, we presented a new and practical implementation of GAs for a {\it SM-GA} as a new symmetric Dual Coding strategy. In this purpose, we tried to improve bounds on GAs convergence by profiting from the manner of operating simultaneously two codings in two units of work to consume the majority of possible representations that can obtained by the two codifications. In this paper, SC and GC were applied to the new proposals. Although, any other coding types and any number of codings could be applied to the sequential and parallel strategies.

Experiments were performed to search for the optimal proposal for a given set of minimization problems. Finding an appropriate best proposition is not an easy task, since each proposal has particular parameters and specific criteria so that the characteristics and typical combination of properties represented by any suggestion do not allow for generalized performance statements. In order to facilitate an empirical comparison of the performance of each proposal, we have measured the success rate progress over generations which transfers a clear view and permits a legal opinion and decision about the efficiency and the evolution of each proposition.

For Serial Dual Coding proposals, Table \ref{tabRes} introduces not bad results according to SR evaluation. Likewise, Figures positioned at the left side of Figure \ref{figSR} prove that each of these proposals enhanced a little the performance of the SGA for a given problem. At least, we can say that they produced results which were best from the worst of those of executing an SGA with unchangeable representation. Thus, it means that their performances maybe were affected by attributing inexact values to their specific parameters, or probably they were affected by the choice of the initial population because this criteria's effect is sometimes dramatic.

As well, in Table \ref{tabRes}, {\it SM-GA} produced relative high results than the other algorithms according to SR measurement and this for all examined functions. The experimental data in this table also suggest that, while it is possible to each proposal to control accurately its parameters, very good performance can be obtained with a varying range of SGA control parameter settings. Figures positioned at the right side of Figure \ref{figSR} show, for each exploited function, a comparison between {\it SM-GA} and SGA referring to SR activity across generations. In these figures, {\it SM-GA} graphical records illustrate how SR was progressing quickly after a small number of generations which made SGA performs better and improves its processing during the investigation for the optimum. On the other side, {\it SM-GA} shows its advancement over {\it SM-GA}$_{SC}$ and {\it SM-GA}$_{GC}$ which proves distinctly the efficacity of blending and integrating two various representations simultaneously.

Experimental results were confirmed by using the t-test results in Table \ref{tabttRes}. Entering a t-table at $398$ degrees of freedom ($199$ for $n_1$ + $199$ for $n_2$) for a level of significance of $95\%$ ($p = 0.05$) we found a tabulated t-value of $1.96$, going up to a higher level of significance of $99\%$ ($p = 0.01$) we detected a tabulated t-value of $2.58$. And to a greater extent, we increased the level of significance to the most higher level of $99.9\%$ ($p = 0.001$) we got a tabulated t-value of $3.29$. Calculated t-test values in Table \ref{tabttRes} exceeded these in most cases, so the difference between compared proposals averages is highly significant. Clearly, {\it SM-GA}$_{SG}$ produced significantly finer results than those of other algorithms by the fact that coexistence of dual chromosomal encryption stimulated production, multiplication and interchange of new structures concurrently between synchronized populations according to the split-and-merge life cycle and ordered functionality.

In futur works, we will use multi-coding {\it SM-GA} in the field of genetic programming, where it exists more enhanced GAs (Evoltution Strategies, state-of-the-art GAs, etc.) and different kinds of representations (tree, linear, etc.), in order to use the well-adapted representation for a specific problem.

Finally, these measurements leave us with valuable perceptions concerning the utility of compounding various coding types for individual representation in collaboration in one SGA. In purpose to ameliorate our new algorithms, we need to have a deeper apprehension of what GAs are really processing as they operate and we are due to understand advantageously and refine our knowing about the specifications of each coding and its reactions with the genetic operators which can help to enhance GAs optimal performances and provide us with more steps towards GAs evolution.

\end{document}